\def\year{2019}\relax
\documentclass[letterpaper]{article} 
\usepackage{aaai19}  
\usepackage{times}  
\usepackage{helvet} 
\usepackage{courier}  
\usepackage[hyphens]{url}  
\usepackage{graphicx} 
\usepackage{graphicx}  
\usepackage{lipsum} 
\usepackage{mathtools}
\usepackage{amsmath} 
\usepackage{amssymb}  
\usepackage{algorithm}

\usepackage{bigdelim}
\usepackage{booktabs}
\usepackage{graphicx}
\usepackage{gensymb}
\usepackage{multirow}
\usepackage{todonotes}
\usepackage{url}
\usepackage{breqn}
\usepackage{subcaption}
\usepackage{amsfonts}
\usepackage{xcolor}
\usepackage{array,multirow}
\usepackage{soul}
\usepackage[normalem]{ulem}
\newcommand{\stkout}[1]{\ifmmode\text{\sout{\ensuremath{#1}}}\else\sout{#1}\fi}
\newcommand{\I}{\mathbf{I}}
\newcommand{\E}{\mathbf{E}}

\newcommand{\X}{\mathcal X}
\newcommand{\M}{\mathbf M}
\newcommand{\Ma}{\mathbf M_{alive}}
\newcommand{\Md}{\mathbf M_{dis}}
\newcommand{\Si}{\mathbb S_i}
\newcommand\tab[1][1pc]{\hspace*{#1}}

\usepackage{xcolor}
\usepackage{xparse}

\title{An Alert-Generation Framework for Improving Resiliency in Human-Supervised, Multi-Agent Teams }
\author{Sarah Al-Hussaini\textsuperscript{\rm 1}, Jason M. Gregory\textsuperscript{\rm 2}, Shaurya Shriyam\textsuperscript{\rm 1}, Satyandra K. Gupta\textsuperscript{\rm 1}\\
\textsuperscript{\rm 1}Viterbi School of Engineering,
    University of Southern California, CA, USA\\ 
\{salhussa, shriyam, guptask\}@usc.edu\\
\textsuperscript{\rm 2}US CCDC Army Research Laboratory, Adelphi, MD, USA\\ 
jason.m.gregory1.civ@mail.mil 
}

\begin{document}

\maketitle

\begin{abstract}
Human-supervision in multi-agent teams is a critical requirement to ensure that the decision-maker's risk preferences are utilized to assign tasks to robots. In stressful complex missions that pose risk to human health and life, such as  humanitarian-assistance and disaster-relief missions, human mistakes or delays in tasking robots can adversely affect the mission.  To assist human decision making in such missions, we present an alert-generation framework capable of detecting various modes of potential failure or performance degradation. We demonstrate that our framework, based on state machine simulation and formal methods, offers probabilistic modeling to estimate the likelihood of unfavorable events. We introduce smart simulation that offers a computationally-efficient way of detecting low-probability situations compared to standard Monte-Carlo simulations. Moreover, for certain class of problems, our inference-based method can 
provide guarantees on correctly detecting task failures.
\end{abstract}

\section{Introduction}
\label{sec:intro}



With the advancement of robotic systems and artificial intelligence, there is a growing interest in more-intelligent, multi-agent teams working collaboratively to accomplish missions. These teams show especially great promise in dull, dirty, and dangerous applications, such as military operations and humanitarian-assistance and disaster-relief (HA/DR) efforts \cite{Gregory2016c}. Despite the widespread use and ever-increasing capabilities of robotic systems, researchers anticipate that human team members will continue to be necessary - and not be replaced by technology - because of various advantages, including diverse expertise, adaptive decision-making, and the potential for synergy \cite{Decostanza2018}. More importantly, HA/DR missions involve tasks with literal life-or-death consequences and so human-in-the-loop operations are mandatory to ensure proper management of resources and critical decision-making authority. Because multi-agent systems will require extensive collaboration, it is imperative that systems be developed with both the strengths and weaknesses of humans in mind, just as researchers and engineers design for the robotic agent. 
The disaster response system needs to integrate humans, robots, and the software agents for effective response to emergency situations \cite{ramchurn2016disaster}. 
Connecting human language \cite{chai2016collaborative}, adapting to human intents \cite{levine2014concurrent}, and co-development of joint strategies \cite{ramakrishnan2017perturbation} can assist successful human-robot teaming.


It is well-understood that the relevant missions in the context of HA/DR are extremely complex \cite{murphy2014disaster}. Due to the size and unstructured characteristics of the operational environment, restrictive communication limitations, and the constant threat of system and task-level failure, there is a large amount of uncertainty in the availability and efficiency of agents
The dynamic conditions introduced by the environment lead to intermittent data flow and require that the agents, including the humans, constantly adapt using the currently-available information. Humans are specifically prone to making mistakes and must overcome cognitively- and emotionally-fatiguing situations in the stressful situations of HA/DR missions \cite{murphy2004human}. This can lead to the issuing of erroneous, slow-paced, or ill-advised commands. In an effort to prevent dangerous scenarios and the catastrophic degradation of performance, the team must be \emph{resilient} to these various challenges, and the accompanying contingencies \cite{shriyam2018incorporating}, by minimizing the impact of human-made errors.

\begin{figure*}[!h]
  \centering
  \includegraphics[width=1.9\columnwidth]{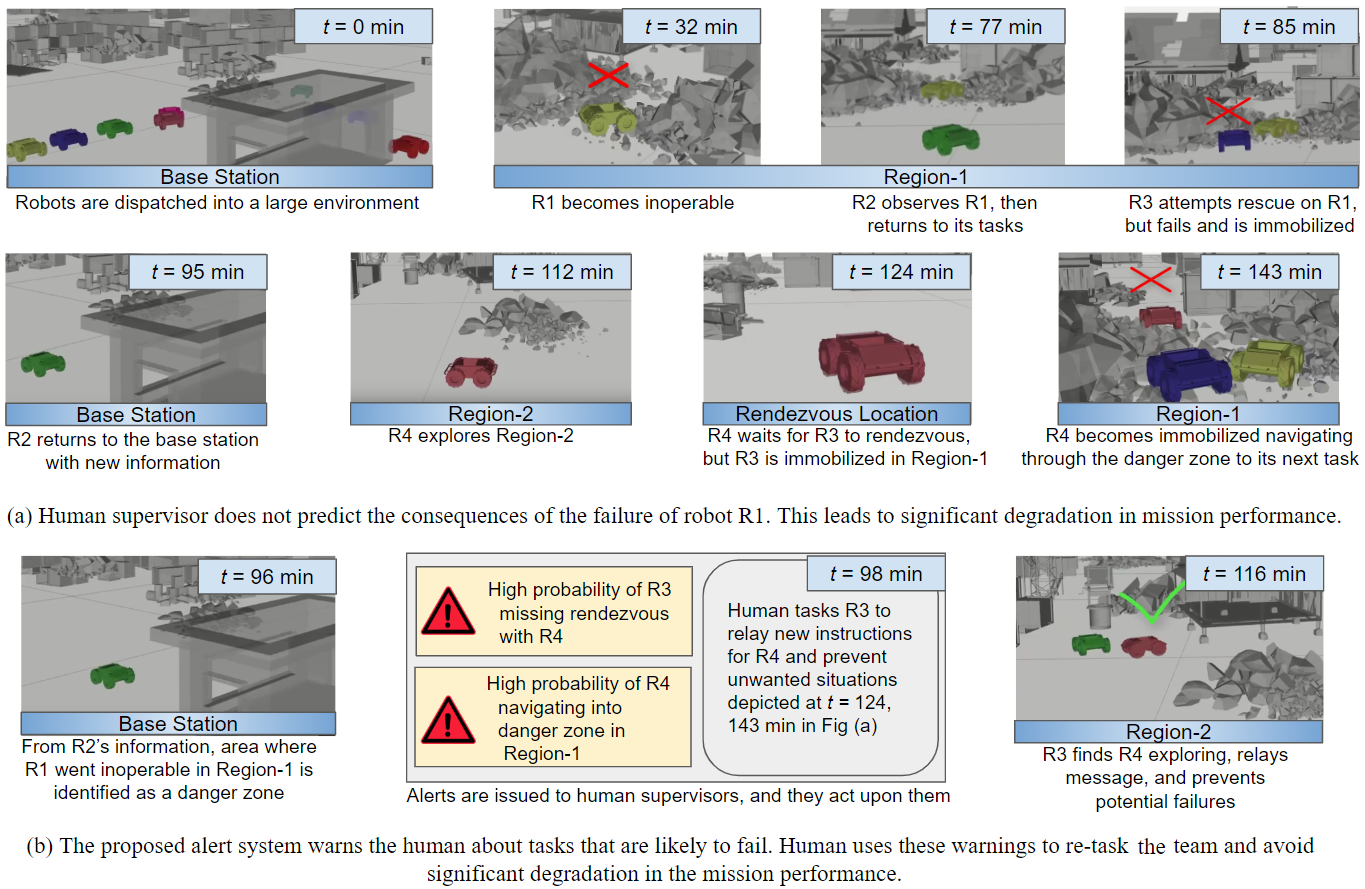}
  \caption{A motivating example scenario with four robots, R1 (yellow), R2 (green), R3 (blue), R4 (brown), visualized in Rviz. }
  
  
 
  \label{fig:scenario}
\end{figure*}
Alerts provide an efficient and effective way for enabling and improving resiliency because they offer a  means to prevent human-introduced mistakes and expedite decision-making. Already, alert systems have been deployed in a number of technologies for every-day tasks, like blind spot detection or lane departure warning systems for drivers to prevent vehicular accidents. An intelligent alert system can also provide tremendous benefit to human-robot teams operating in time-sensitive, safety-critical scenarios. This is because an alert system can assist the team with the required decision-making that is inherently challenging, taxing, and severely-consequential. Already, researchers have investigated the human factor concerns associated with supervisory control of multi-robot systems \cite{wong2017workload}, \cite{sherwood2018effect},  \cite{chien2018attention} and there exists several different alert-generating architectures and interfaces such as an augmented reality-based solution \cite{makris2016augmented} and the various alert systems used by NASA \cite{mosier2017autonomous}. There have also been several systems designed specifically for human-robot teams in the HA/DR context, both with 
\cite{barnes2014designing}  \cite{jentsch2016human} and without alert systems \cite{kruijff2014experience}. These alert-generation frameworks are purely reactive, where a notification is provided to the human once an undesirable event has occurred. The most relevant work to date is the predictive conventional interface and predictive virtual reality interface designs \cite{roldan2017multi}. These two interfaces predict the risk and relevance of a robot performing some task. In this work, we also seek a proactive approach to providing alerts; however, our focus is on the generation of alerts for potential dangers or unwanted situations in the mission, as well as, erroneous and inefficient task assignments issued by humans. The Human Factors Analysis Classification System framework, applied for human operators in different applications \cite{HFACS-aviation-2007}, \cite{HFACS-medical-2014}, identifies decision errors, perceptual errors, and violations as the triggers to unsafe acts, which have preconditions from environmental, personnel factors, and mental and physical condition of operators. Our alert system is aimed to address all these factors, and improve the performance of human supervisors. A motivating mission scenario is illustrated in Figure \ref{fig:scenario}.


Building a useful alert system suitable for multi-agent systems requires careful consideration. First, a proper language-based scheme is necessary to facilitate intuitive interaction with the human teammates.  Unlike the aforementioned alert systems for every-day tasks, an alert system for multi-agent systems cannot simply rely on sensors and comprehensive observations. Instead, a fieldable alert system requires inference and probabilistic model estimation to account for the inherent uncertainty in mission operations. An alert system must also be flexible to the types of alerts offered, be tailored to the human preferences and mission needs, and generate meaningful alerts in a timely fashion so that agents can take the necessary, corrective actions. In this work, we propose a novel, alert-generation framework that overcomes these challenges to improve resiliency of multi-agent teaming. This work provides two noteworthy contributions. First, we define a formal language and a state machine-based simulation architecture to model the likelihood of salient system states during the execution of a human-commanded mission. Second, we present an inference engine that compares human-specified alert conditions with the probabilistic outcomes of the state machine simulations to produce worthwhile alerts. Using a probabilistic temporal logic framework, we enable the human to specify alert conditions based on their requirements and preferences in an effort to improve the value of the reported alerts. To demonstrate the usefulness of our proposed framework, we provide some example scenarios to be detected as results. They show the detection of unwanted situations based on new information which can be complex for humans to infer by themselves in time-critical missions. We also demonstrate the usefulness of smart simulation which can be useful in detecting low probability events in computationally efficient manner compared to Monte-Carlo simulations.



\section {Overview of Approach}
\label{sec:overview}

\subsection {Mission Description}
\label {sec:mission}

We consider a disaster-stricken environment, e.g., a city after an earthquake, flood, or wildfire, as a representative environment for a generic HA/DR mission. The outdoor environment is assumed to be on the order of several square kilometers and comprised of complex, unstructured terrain. A human-supervised team of robots is deployed for exploring affected regions efficiently, collecting important information, and performing certain tasks, according to human preferences. As the team of robots navigate through the environment, there is some nonzero probability of operational failure, which could be a result of spatial factors, such as complex terrain, or stochastic events, such as hardware or software failures.
There also are communication challenges because of the large scale operational environment. Limited communications cause substantial delays in humans receiving information from the robots in the field.
We encode these typical challenges and characteristics of a generic HA/DR operation in our specific mission in order to present and test our proposed alert generation framework. Thus, any other multi-robot mission, related to HA/DR-relevant application might be reduced to a variation of our defined mission, and our framework can be tailored for any such operation.

In a large-scale environment, usually there are some regions of higher importance which should be given priority in the exploration process. We assume that the humans, typically the first responders, can use their expertise and protocols to identify some \textit{areas-of-interest} (AoIs), chosen at the beginning of the mission, or dynamically through out based on latest information. We also assume, there are a few, very sparsely located beacons, and robots need to be physically close to them in order to communicate with humans. Each robot carries out several tasks based on its observations and the situation-specific instructions, and then navigates back to a beacon after some time to reconnect with its supervisors. 

In our mission, human supervisors at the base station provide instruction-set to each robot in the team, and dispatch them for exploration and data collection. In addition to navigation and exploration, humans can instruct a robot to rendezvous with another robot, or relay new instructions to another robot. Humans can also command a robot to provide assistance to a temporarily-disabled robot in an attempt to make it functional again. Every time humans receive new information from a returning robot, they need to assess the current situation, and make intelligent decisions based on the latest update. Humans might want to re-task the fielded robots, which may be currently out of communications range, in order to avoid unwanted situations and prevent undesirable contingencies. Thus, the human supervisors are under constant pressure to make decisions quickly in order to utilize the robots most effectively. Due to the large size of the environment, humans may only receive updates from each robot after some extended time. Therefore, the robots need to be sufficiently tasked for prolonged periods, and the stochasticity and danger involved in the mission may prompt the humans to give instruction-sets which are incredibly complex. This necessitates a systematic way for humans to command robots, and deployed systems should be equipped to describe a complete instruction-set with adequate complexity.

\subsection{Language For Commanding Robots}
\label{sec:language}

We present a formal language in this section that can decompose the complex, human-provided instructions in a systematic way. First, we define a set of tasks that a robot can execute; these are: explore, navigate, rescue, rendezvous, relay, return, wait. 
The uncertainties in the mission, and different stochastic phenomena demand several other actions from the robots, in addition to exploration and navigation. In order to improve resiliency in a mission with a high probability of failure, robots need to provide assistance to temporarily-disabled teammates by attempting \textit{rescues}, as defined in our previous works \cite{Al-Hussaini_2018_IROS}, \cite{Al-Hussaini_2018_SSRR}. This rescue operation has stochastic outcomes, such as successful rescue, failed rescue, and, in the worst case, the rescuer robot also becoming inoperable during the attempt. In order to tackle the challenge of scarce communication in a HA/DR mission, we have included \textit{rendezvous} and \textit{relay} tasks. In a collaborative exploration-based mission, it is useful for multiple robots from different regions to meet at pre-scheduled times and locations to exchange information, referred to as \textit{rendezvous}. 
The \textit{relay} task requires a robot to go to a specific region, search for another robot, and relay a specific piece of information to that robot. Here, we have limited the scope of relay tasks to initially clear the other robot's old instructions and then issue a new command-set. 
Since one of the focuses of this work is to provide alerts for future contingencies, this \textit{relay} task is particularly helpful. It can be used in sending an available robot to prevent an adverse situation happening to a robot already in the field. 
Finally, the return and wait tasks correspond to the robot navigating back to within communications range of the human supervisor, or remaining stationary in one location, respectively.

{\renewcommand{\arraystretch}{1.2}
\begin{table*}[!h]
   \scriptsize
   \centering
   \caption{Supporting Functions for Robot Instructions and Alert Conditions} 
  \begin{tabular}{l|l}
  \toprule
   \textbf{Functions or Literals} & \textbf{Description of Return Variables or Values} \\ 
  
    \hline
    
    $TravelDuration_i(\X_1,\X_2)$
    &
    Estimated travel robot $R_i$ to navigate from region $\X_1$ to  $\X_2$
    \\
    \hline

    $isRiskyRegion(\X)$
    &
    $True$ when the region $\X$ is risky, $False$ otherwise
    \\
    \hline
    
    $NearbyRobotID_i (st,d) $
    & ID of robot with state type $st$, and within distance $d$ from robot $R_i$
    \\
    
    \hline
    
        $CountNearbyRobots_i (st,d) $
    & Number of robots with state type $st$ , within distance $d$ from $R_i$
    \\
    
    \hline
    
    $CountExploringRobots (\X) $
    & Number of robots exploring region $\X$
    \\
    
    \hline
    
    $CountEvents (e, j_1, a_1, j_2, a_2, ..., i_m, a_m)$
    & Number of events in $\mathbf E_{R_i}$ whose $type$ is $e$, and  $arg^{j_1}, arg^{j_2}, ..., $
    \\
    &  $arg^{j_m}$ are respectively $a_1, a_2, ..., a_m$ \\
    
    \hline
    
    $ToRend_i =\left\{
  \begin{array}{@{}ll@{}}
    1, & \text{if } S_{R_i}.type=TravelToRend   \\
    0, & \text{otherwise}
  \end{array}\right.
    $
    & $True$ if robot $R_i$ is travelling to rendezvous location
   \\
   
   \hline
   
      $EndRend_i =\left\{
  \begin{array}{@{}ll@{}}
      1, & \text{if }   (\, StateHist_{R_i}^1.type=WaitToRend,    \\
    & S_{R_i}.type \neq WaitToRend\,) \text{ or } S_{R_i}.type=Rend\\
    0, & \text{otherwise}
  \end{array}\right.
    $
    & $True$ if robot $R_i$ moves to a new task after a rendezvous attempt
   \\

   \hline

   $ IsNav_i =\left\{
  \begin{array}{@{}ll@{}}
    1, & \text{if }  S_{R_i}.type \in \{ Navigating, TravelToExpl,  \\
    & \quad  TravelToResc, TravelToRel, TravelToRend \}   \\
    0, & \text{otherwise}
  \end{array}\right.
    $
    & $True$ if robot $R_i$ is navigating or travelling to a task location
   \\
   
   \hline
    
    $NeverRescue (i, \mathbf J)=\left\{
  \begin{array}{@{}ll@{}}
    1, & \text{if } \forall s \in \mathbb S_i^*, \; \lnot( s.type = rescue, \, s.arg^1 \in  \mathbf J ) \\
    0, & \text{otherwise}
  \end{array}\right.
    $
    & $True$ if robot $R_i$ will never attempt rescue on a robot with ID $\in \mathbf J$
   \\
   
   \hline
    
      $NeverRelay (i, \mathbf J)=\left\{
  \begin{array}{@{}ll@{}}
    1, & \text{if } \forall s \in \mathbb S_i^*, \; \lnot ( s.type = relay, \, s.arg^1 \in  \mathbf J)  \\
    0, & \text{otherwise}
  \end{array}\right.
    $
    & $True$ if robot $R_i$ will never attempt relay on a robot with ID $\in \mathbf J$
   \\
   
   \hline

    \addlinespace[1.3ex]
        $MinTravelT(i,L) =\dfrac{EuclideanDistance(L_i, L)}{MaxNavigationSpeed_{R_i}}$
    & Minimum navigation time for $R_i$ from its location $L_i$ to location $L$
   \\
    \addlinespace[1.3ex]
   
   \hline     
\multicolumn{2}{c}{$^*$ Defined in Inference-based Approach Section }\\

  \end{tabular}
  \label{tab:supporting_func}
\end{table*}
}

\begin{figure*}[!h]
  \centering
  \includegraphics[width=2\columnwidth]{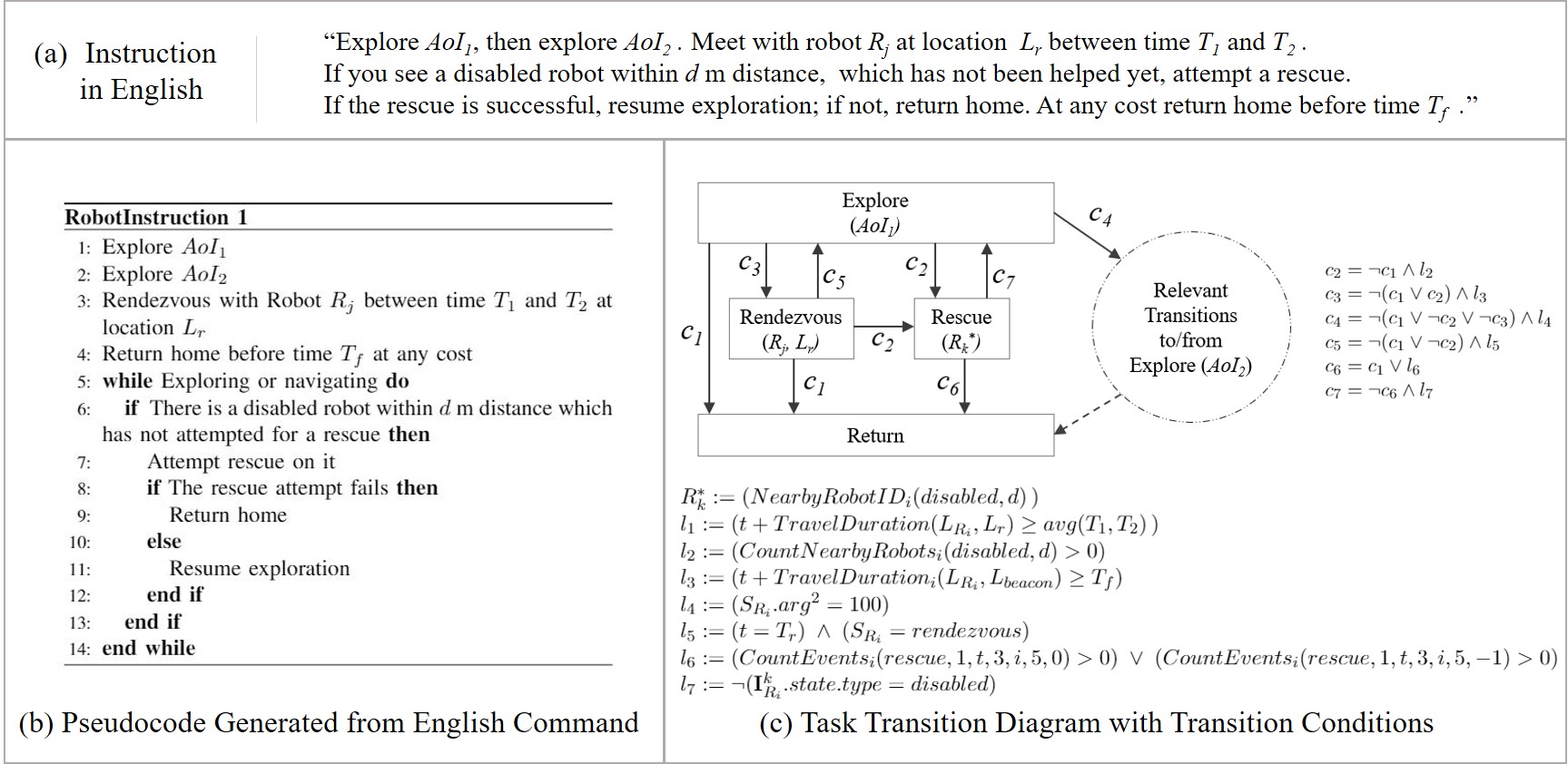}
  \caption{An example of the language decomposition offered by our proposed framework. English instructions provided by the human (a) are converted to pseudocode (b) that is used to generate a task transition model (c).} 
  \label{fig:instruction}
\end{figure*}

Let $\mathbf{R} = \{R_1, R_2, ... , R_N\}$ be the set of $N$ robots, deployed in the mission. At any time $t$, each robot $R_i$ has state $S_{R_i}$, location $L_{R_i}$, event list $\mathbf E_{R_i}$, list of other robot's most recent status updates (location, state) $\I_{R_i}$, and its own state history list $StateHist_{R_i}$. 
A robot goes through a series of states in order to perform a particular task. For example, \textit {TravelToRend, WaitToRend, Rendezvouzing} are the states corresponding to task \textit{rendezvous}; all having the same argument-list, i.e., robot ID to meet, rendezvous location, and time window. 
The element $StateHist_{R_i}^\tau$ gives its state from a time that was  $\tau$ instances before the current time. A complete list of events should include relevant environmental events as well as some task related events  with a robot itself or other teammates. An example event type is \textit{rescue attempt}, with arguments robots IDs, location, time, and outcome. Every unique event a robot $R_i$ learns about is stored in $\mathbf \E_{R_i}$. In order to create time-based and situation-dependent instructions, humans need to construct different conditional statements which make use of time, the robot's information state at the instance, and model estimation of different parts of the system. For crafting these conditions, we present a compiled list of functions in Table \ref{tab:supporting_func}. We use items from this list for the robot's instruction-sets and alert conditions presented in Figure \ref{fig:instruction}, and Tables \ref{tab:warning conditions}, \ref{tab:inference}, respectively.


Humans issue complex instruction-sets to the robotics teammates. Each instruction-set is defined as a set of tasks, arguments, event- and temporal-based conditionals. Based on the different outcomes of the stochastic parameters related to the environment, events, and the states of other teammates, a robot may perform various sequences of tasks for the human-provided instructions. The first step is to convert the instructions to a pseudocode format in order to properly identify the $if$ and $while$ statements along with the conditions. All the conditional statements of interest in this work can be expressed mathematically with the functions and variables described in this section. Additionally, we identify proper arguments for each task in the instruction-set. These aid with generating a \emph{task transition model} for each robot where each transition is dictated by a condition. We provide an illustrative example of language decomposition for a snippet of an instruction-set in Figure \ref{fig:instruction}.

\subsection{System Architecture}

\begin{figure}[!h]
  \centering
  \includegraphics[width=.99\columnwidth]{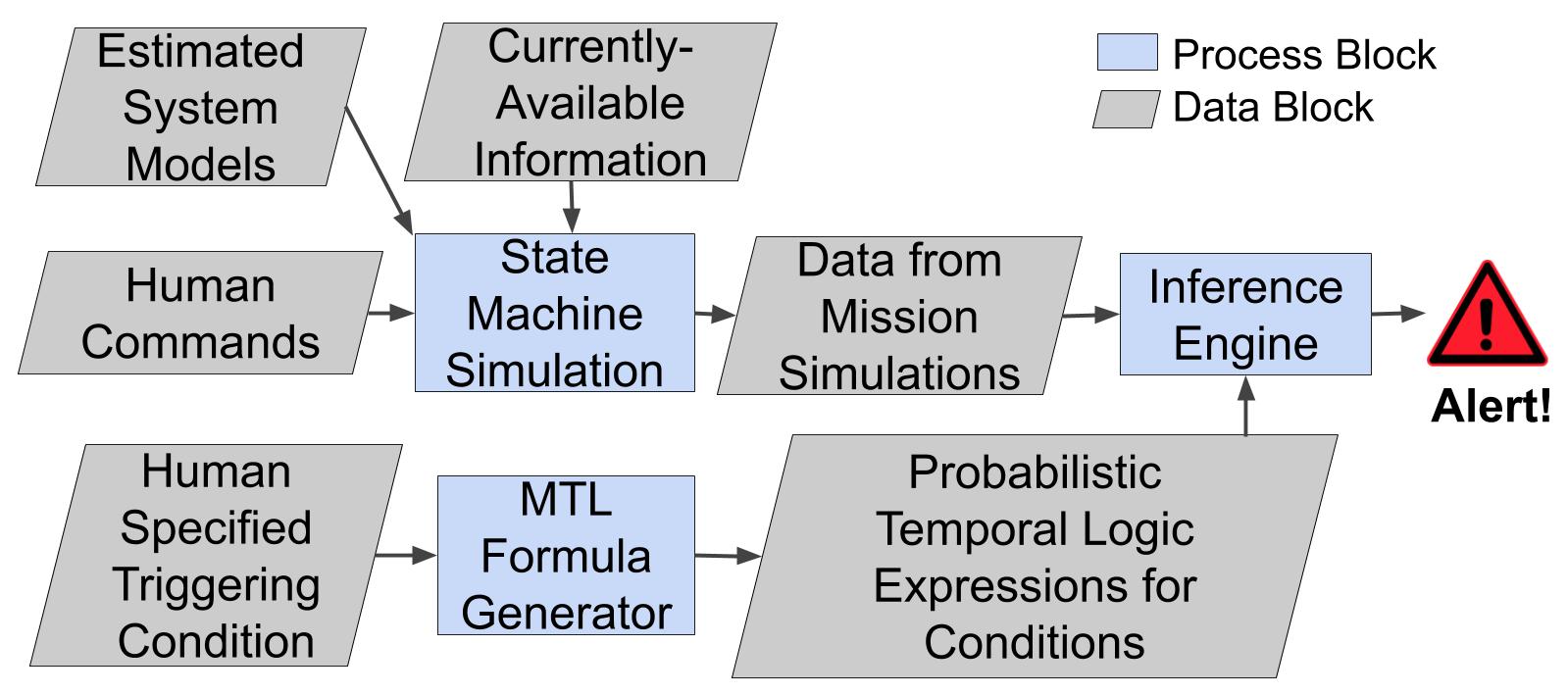}
  \caption{Block diagram of the proposed alert-generation framework. The grey slanted rectacles are data blocks and blue rectangles are processing components of the system.
  \label{fig:block}} 
\end{figure}

Our proposed framework is targeted to be used by humans supervising a team of robots in a challenging large-scale mission. The task transition model derived from complex instruction-set is used to model each robot's behavior as a state machine. Whenever any robot returns to a human with new information, they provide state information of other robot teammates encountered in the field. We assume there are some estimated models of different stochastic parameters within the entire system. Using these models, the latest state information, and the state machine models of the robots, forward simulations of the entire system are performed. These simulations are done at appropriately high- level, in order to perform a large number of simulations quickly and generate immediate alerts. Each simulation run is a collection of parallel, but inter-dependent, state machine simulations for all robots in the field. The results of simulations give probabilistic estimates on feasible outcomes in the mission. Simultaneously, humans define their preferred list of unwanted situations that they feel are important to detect. These contingency conditions are then expressed as mathematical propositions. Our framework provides an inference engine that utilizes the results from the simulations, and finds \textit{Truth} values for the user-specified alert conditions. If any of these become $True$, the framework shows the corresponding alerts to the humans. This alert can be based on probabilistic estimates from the simulations, or it can provide guarantees on particular situations happening with $100\%$ certainty. The proposed framework is represented in the block diagram in Figure \ref{fig:block}.

\subsection{Specification of Alert Conditions}
\label{sec:spec}


We have identified some exemplary alert-triggering scenarios that humans may find useful and relevant to an HA/DR mission. We also outline mathematical expressions of the alert conditions for detection of these situations. We formulate the conditions in a probabilistic temporal logic framework \cite{probabilistic_temporal_13}, using different parameters and functions. 
Probabilistic temporal logic is a powerful language to mathematically express many kinds of complex conditions. Humans are free to choose different alert conditions from a potential list relevant to each mission, or craft their own conditions based on their preferences. 
The description of several alert situations are enumerated in this section, and the mathematical expressions relevant to these contingencies are provided. Note, this is not a comprehensive list for a mission, rather we provide some worthwhile examples depicting different types of fundamental expressions and conditions. 

In our proposed framework we use Metric Temporal Logic (MTL) \cite{MTL_08} specifications to detect a particular unwanted situation in a single mission outcome.
Mission progression ends at some time $T_{end}$ once all of the robots are in \textit{returned} or \textit{disabled} states; after that time we can assume the entire system remains constant. 
Let $\Psi$ be a set of atomic propositions, crafted from the aforementioned items relevant to the mission threads, and the MTL formulae are built from $\Psi$ using Boolean connectives ($and$ $\wedge$, $or$ $\vee$, $not$ $\lnot$), propositions $\top (True)$ and $\perp (False)$, and time-constrained or -unconstrained versions of temporal operators ($eventually\, \lozenge, \, always \, \square, \, next \, \bigcirc, \, until \; \mathcal U$). A time-constrained temporal operator is $\Gamma_I$, where $\Gamma \in \{ \lozenge, \square, \mathcal U \}$, and time interval $I\subseteq (0,\infty)$, while the unconstrained version is $\Gamma \equiv \Gamma_{(0,\infty)}$. If we generate a large number of probable strings of mission threads, representing the different ways the mission can progress, we can compute the fraction of strings that satisfies an MTL formula. This fraction can be taken as the estimated probability which is compared with a threshold value $p_{th}$ in order to detect high or low probability conditions. $\mathbf P_{\sim p_{th}} \Phi$ indicates that the probability of $\Phi$ being $True$ is $\sim p_{th}$, where, $\sim \in\{<,\leq,>,\geq,=\}$,  $0\leq p_{th} \leq 1$, and $\Phi$ is an MTL formula. For clarity, we provide some examples of MTL formula similar to the ones used in Table \ref{tab:warning conditions}, along with their implications in words.

\begin{itemize}
    \item $(\lozenge_{(t_1,t_2)} \phi)$ is $True$ \textbf{\textit{iff}} eventually at some time between $t_1$ and $t_2$, $\phi$ is $True$
    \item $(\lozenge ( \phi_1 \rightarrow \bigcirc \phi_2) )$ is $True$ \textbf{\textit{iff}} eventually at some time $\phi_1$ is $True$ and right after that (at the next time instance), $\phi_2$ is $True$ .
    \item $(\lozenge ( \phi_1 \rightarrow (\phi_2 \, \mathcal U_{\{t'\}} \top ) \,)$ is $True$ \textbf{\textit{iff}}
    eventually at some time $\phi$ is $True$, and right after that $\phi_2$ remains $True$ for next $t'-1$ time instances.
\end{itemize}

{\renewcommand{\arraystretch}{1.2}
\begin{table*}[!h]
   \scriptsize
   \centering
   \caption{ Example Alert Condition Specifications: \textit{Probabilistic estimation for enumerated  situations} }

  \begin{tabular}{c|l|l}
  \toprule
    \textbf{Alert \#} & \multicolumn{1}{c}{\textbf{Condition descriptions} } & \multicolumn{1}{c}{\textbf{Expressions of conditions} }\\ \hline
    
    
    & Non-zero probability of robot $R_i$ having states $S_1, S_2, S_1, S_2$
    & $\mathbf {P}_{> 0} [
    \lozenge ((S_{R_i}=S_1)\rightarrow \bigcirc( S_{R_i}=S_2)\rightarrow \bigcirc (S_{R_i} = S_1) $ \\ 
    
    1 &  respectively in four consecutive time steps
    &\tab \tab $\rightarrow \bigcirc (S_{R_i} = S_2)]$ \\
    
    &  (oscillation between states $S_1,S_2$) 
    &$\text{Where, } (\{S_1,S_2\} \subset \mathbb S)  \; \wedge \; (S_1 \neq S_2)$ \\
    
    \hline
    
    
    &  (i) High robability of robot $R_i$ and $R_j$ never being together at the 
    &   (i)    $ \mathbf {P}_{\geq p_{th}} [
    \lozenge_{(t_1,t_2)}\, \lnot \, (L_{R_i}=X\, \wedge \, L_{R_j}=X))
     ] $ \\ 
     
     2  & scheduled location ($X$) within time $(t_1,t_2)$ 
     &  (ii)    $ \mathbf {P}_{\geq p_{th}'} [
    \, B_{j} \, 
    \vee B_{i} \,
     ] $\\
     
     &  (ii) High probability of one of the robots $R_i, R_j$ being at scheduled 
     &Where,
       $  B_{i} := (\lozenge_{(t_1,t_2)}L_{R_j}=X) \, \wedge \,  (\square_{(t_1,t_2)} \lnot (L_{R_i}=X))  $\\
       
      &  time \&  location while other one does not
      & $  B_{j} := (\lozenge_{(t_1,t_2)}L_{R_i}=X) \, \wedge \,  (\lozenge_{(t_1,t_2)}\square \lnot (L_{R_j}=X))  $ \\ 
       
       \hline

    
        & (i) Low probability of robot $R_i$ attempting rescue on $R_j$ at location $X$
       &  (i) $\mathbf {P}_{\leq p_{th}}  \, [
    \lozenge \; (\, CountEvents_i (rescue, 2, X, 3,i, 4,j)>0\, )
    ] $ \\ 
    
   3 & (ii) Non-zero probability of robot $R_i$ attempting rescue on $R_j$ 
       &  (ii) $\mathbf {P}_{> 0}   \, [
    \lozenge\; (\, CountEvents_i (rescue, 2, X, 3,i, 4,j)>n\, )
    ] $\\
    
    & at location $X$ more than $n$ times
    &  \\

    \hline
    
    
    4 & High probability of robot $R_i$ navigating through a risky region
    &   $ \mathbf {P}_{\geq p_{th}}  \, [
    \lozenge \, isRiskyRegion (L_{R_i})
    ]$ \\ 
    \hline
    
    
    & High probability of having more than $N_Y$ number of robots exploring
    &   $\mathbf {P}_{\geq p_{th}}  \, [
     \lozenge ( \; \varphi_Y \rightarrow (\varphi_Y \; \mathcal U_{\{t'\}} \top \,) \; )
         ]$ \\ 
     
     5 & region $Y$ at once (more than a time duration t'), where $N_Y$
    &  Where, $\varphi_Y := (\, CountExploringRobots (Y) \geq \, N_Y)\    $ \\ 
     
      & is an upper bound on number of exploring robots for that region
    &  \\

     \hline
     
     
     & (i) High probability of $R_i$ travelling to rendezvous for 
    &  (i) $\mathbf {P}_{\geq p_{th}}  \, [
     \lozenge ( \;ToRend_{R_i} \rightarrow (\, ToRend_{R_i} \; \mathcal U_{\{t'\}} \top \, ) \; ) 
         ]$ \\ 
     
     & more than a   time duration $t'$  
    &  (ii)$\mathbf {P}_{\geq p_{th}}  \, [
     \lozenge ( \;EndRend_{i} \rightarrow \, 
     (\, IsNav_{i} \; \mathcal U_{\{t'+1\}} \top \, ) \; ) 
         ]$ \\
     
   6   & (ii) High probability of $R_i$ navigating for more than  $t'$ time duration
    & 
    \\ 
    & towards the next task, after a successful or failed rendezvous 
    &  \\

    \hline
     
  \end{tabular}
  \label{tab:warning conditions}
\end{table*}
}

The following items narrate some contingency situations that human supervisors may want to receive alerts for. The detection condition corresponding to each situation and its mathematical expression are provided in Table \ref{tab:warning conditions}.  How to choose appropriate probability threshold values, in accordance with human preferences on certain situations in a mission, is to be considered for our future work.

\begin{enumerate}
    \item \emph{Wastage of time due to oscillatory states}
    
    Poorly-defined conditionals in an instruction-set can cause a robot to oscillate between two states 
    which might not be intended by the human commander. This may waste a significant amount of time 
    without any progress.

    \item \emph{Improbable rendezvous}
    
    A pre-scheduled rendezvous between two robots might not happen if humans make mistakes in issuing the commands. 
    It can also be missed if 
    a robot
    becomes inoperable, or skips the rendezvous 
    to execute a separate branch of tasks due to situations and specifications. In crafting the condition expression, we assume that both robots arriving at the designated location within the scheduled time period is sufficient for a successful rendezvous.

    \item \emph{Improbable or redundant rescue attempts}
    
    A specific rescue task instructed to a robot may never actually occur due to certain, obstructing events. Another unwanted situation can be where a robot continuously attempts multiple rescues of the same robot, when it is not intended behavior.
    Incomplete or improper conditional statements tied to the rescue task can be a possible reason for this issue.
    In a broader sense, redundant rescue might also refer to multiple robots attempting the same rescue.

    \item \emph{Navigation through  a risky region}
    
    If humans receive information about a newly-assessed, risky region they might want to check whether any robot in the field is likely to navigate through that region based on its instructions. Moreover, humans can  erroneously assign a robot to a risky region. If humans are notified of this condition in a timely fashion, dangers can be prevented.

    \item \emph{Redundant exploration}
    
    Every region, based on its characteristics, has an optimal number of robots for achieving fast, collaborative exploration. If there are too many robots exploring a relatively small region together, it may make the exploration process inefficient. We have provided an expression for such conditions in Table \ref{tab:warning conditions}. Likewise, redundant exploration may occur when a robot is exploring a region that has already been explored by another robot.
    

     \item \emph{Excessive travel time for rendezvous or rescue}
     
      Humans might want to avoid having rendezvous or rescue at a location where a robot needs to navigate for a long time from its previous task  or to its next task. We have provided expressions for rendezvous in Table \ref{tab:warning conditions}.
     
\end {enumerate}

There can be many other alert conditions that humans may want to detect, such as, risky rescue, long travel-time for an unsuccessful task, etc. Any condition of a person's choice can be mathematically formulated, used for detection of adverse situations, and produce an alert for the humans.



\subsection{Methods to Detect Alert Conditions}

Given the specification of alert conditions, the alert system must identify when an alert condition is expected to hold $True$, and issue an alert to the humans. The alert condition detection can be simulation- or inference-based. Our simulation-based approach issues alerts based on probability estimates from simulations, but it lacks any guarantee. On the other hand, the  inference-based approach may not be possible in every situation, but when applicable, it is able to provide a guarantee on certain situation.

\subsubsection{Simulation-based Approach}

We use high-level, discrete-time simulations of the mission, where each simulation uses the system model and the latest information. It generates data on a single instantiation that the system could progress throughout the mission, out of infinitely many possibilities. We perform a large number of simulations, and observe what percentage of the simulations have a specific, unwanted situation of occurring. This is considered as the estimated probability of the alert condition to hold true. If this probability meets the thresholding specification set by humans, the interface shows an alert about the possible contingency.


If we consider detection of an extremely low-probability situation, 
repeated random sampling or 
Monte-Carlo simulations may become computationally infeasible due to the excessively large number of simulations required. Our smart simulation feature provides a computationally feasible way to detect even these low-probability circumstances. There are often some critical events that prompt the unwanted situation to be detected. Such critical events are identified from the state or task transition conditions, and are artificially triggered during the simulation runs. The probability estimate from these simulations give the conditional probability, and using the probability of the critical event, we estimate the actual probability of the unwanted situation.

\subsubsection{Inference-based Approach}


The estimated probability of alert-triggering situations, calculated from  simulations, might produce a poor representation of the real scenario,  as the uncertainty and size of the system increases. Also, it is difficult to build an adequate model of the stochastic parameters in the mission. Therefore, we attempt to perform quick inferences using only the non-stochastic parts of the system, (i.e., instruction-sets of the robots, maximum navigation speed) instead of using simulations to detect certain contingencies with significantly-higher confidence. For a complicated mission, it might not be always possible; nevertheless, it can produce useful results in some cases. 

Many of the alert conditions that we have presented refer to the probability of certain tasks being completed, i.e., certain states and/or locations that are reached by one or more robots  within a time range. If an alert condition can be modeled in this way, and the simulation-based testing shows absolutely zero probability, this inference-based approach can be attempted. This method firstly converts the state transition model of each robot into a directed graph without the transition conditions. 
Using a graph search algorithm, like Depth First Search (DFS), to conduct reachability tests, and considering some simple Boolean literals, this approach can provide the humans with a much stronger assessment of an alert condition. In fact, in these cases 
it can guarantee absolute certainty. 

{\renewcommand{\arraystretch}{1.2}
\begin{table*}[!h]
   \scriptsize
   \centering
   \caption{ Inference-Based Alert Conditions: \textit{Prove that robot $R_i$ can not reach state $S_f$ and location $L_f$ by time $T_f$} }
  \begin{tabular}{l|l}
  \toprule
     \multicolumn{1}{c}{\textbf{Condition descriptions} } & \multicolumn{1}{c}{\textbf{Expressions of conditions} }\\ 
    \hline
    
   1) $R_i$ is functioning, $S_f$ is not reachable from its current 
   &
   $(i \in \Ma) \; \wedge \;
   (S_f \notin \Si) \; \wedge \;
   ( \; \forall j\in \Ma, \; 
   $
   \\
   
   state,  and no functioning robot will attempt a rescue on 
   &
   $
   ( \; NeverRelay(j,\{i\}) \; \wedge \; 
   NeverRescue (j, \Md) \; ) \;)
   $
   \\
   
   any disabled  robot, and not relay new instruction to $R_i$
   &
   \\
   
   \hline
   
   2) $R_i$ is disabled and no functioning robot will attempt any  
   &
   $(i \in \Md) \; \wedge \;
   ( \; \forall j\in \Ma, \;
   NeverRescue (j, \Md) \; )
   $
   \\
   
   rescue on any robot who are disabled 
   &
   \\

   \hline

   3) $R_i$ is disabled, $S_f$ is not a reachable state even if $R_i$ is  
   &
   $(i \in \Md) \; \wedge \;
  (S_f \notin \Si) \; \wedge \;
   ( \; \forall j\in \M-\{i\}, \;
   NeverRelay (j, \{i\}) \; )
   $
   \\
   
    rescued,  and no robot (functioning, or disabled robot after  
   &
   
   \\
   
   being rescued) will attempt   relaying new instruction to $R_i$
   &
   \\

   \hline
   
  4) $R_i$ is functioning, but its earliest possible arrival time to 
   & $ (i\in \Ma) \; \wedge \;
   (t_i + MinTravelT(i,L_f) > T_f) 
   $
   \\
   
 destination location $L_f$ is later than time $T_f$
   &
   
   \\
   
   \hline
   
  5) $R_i$ is disabled; even if it is  successfully rescued at the  
  &
  $ (i\in \Md) \; \wedge \;
  (EarliestRescueTime(i) + MinTravelT(L_i, L_f) > T_f )
  $
  
   \\
   
 earliest possible time based on other robots' times and 
   &
   where, if $\tau$ is an array with elements $\tau^j = t_j+MinTravelT(j,L_i), $ 
   
   \\
   
   locations,  it still can not reach $L_f$ in time $T_f$
   &
   $\; \forall j \in \M - \{i\}, \quad EarliestRescueTime(i)= max (min (\tau), t_i)$
   \\
   
   \hline

  \end{tabular}
  \label{tab:inference}
\end{table*}
}

It may appear as though confirming reachability for an individual agent may be sufficient in this inference. However, HA/DR missions are actually much more challenging and tasks, such as \textit{rescue} and \textit{relay}, can complicate the inference process. Let, the latest information available to humans on each robot $R_i \in \mathbf R$ is from a time in the past, $t_i$. Its state, previous state before arriving to the latest state, and location are $S_i, S_{prev,i}, L_i$, respectively. We can perform a graph search for each robot $R_i$, and obtain the set of all possible states that $R_i$ can reach from any state $s$. Note, an inoperable robot does not have any reachable state unless it is successfully rescued by another robot, and then it returns back to its previous state. Let, $\mathbb S_i$ represent the set of all reachable states for robot $R_i$, from state $S_i$ if it is functional, and from state $S_{prev,i}$ if it is not functional. Let $\mathbf M_{alive}, \mathbf M_{dis}$, $\mathbf M$ represent the sets of IDs of functional, non-functional, and all robots, respectively. 
Some instruction-set for a robot may require dynamic assignment of an argument of a certain task, for which the state transition model of the robot's corresponding states might have unknown argument values. For example, the instruction can be ``if you see a robot within $20$m, rescue it". If this $rescue$ state is $s$ in the state-graph, {$s.arg^1$} will be `unknown', hence $s.arg^1 \notin M$. It is important to differentiate this unknown argument from known arguments in order to properly assess the probability of rescue or relay operation for a robot. 
Given all the specifications, we would like to prove whether $R_i$ will never reach state $S_f$ at location $L_f$ by time $T_f$. The conditions to be checked for this purpose are given in Table \ref{tab:inference}, using some supporting functions from Table \ref{tab:supporting_func}. The first three conditions of Table \ref{tab:inference} test for reachability to the final state $S_f$, while the last two check whether it is physically possible for $R_i$ to navigate to the destination $L_f$ in time $T_f$.


\section {Results}

We have tested several alert conditions in our Python-based custom simulator for some mission scenarios that capture the inherent challenge for a human to quickly process a complex inference or dependency while observing only raw data. We then visualize the resulting outcomes using Rviz from the Robot Operating System (ROS). We provide a few examples in this section to illustrate the value of our alert system and to demonstrate the applicability of alert condition detection. An example case shows how the inoperability of one robot decreases a second robot's probability of successful rendezvous by between $0-41\%$, depending upon the time and location of the disabled robot. Without an alert system, it might not be possible for the humans to infer such contingencies under pressure, and would not be able to make intelligent decisions that support resilient operations.

\subsection {Using Simulation to Issue Alerts}
\label{sec:simulation}


The individual robots as state machines in the simulation are dependent on each other and the environment. While staying within a state, a robot keeps performing some low-level actions, which affect itself 
and its teammates. At the start of the simulation, each state machine, i.e., robot model, needs to be properly initiated according to latest update. We start simulation from the earliest update time instance among the robots, and forward simulate all of the robots whose statuses are not known for respective time instances.

 \subsubsection{Is there a non-zero probability of robot $R_i$ oscillating between two tasks $T_1$ and $T_2$? }
 Let robot $R_i$ be currently in the field, with the instruction-set described in Figure \ref{fig:instruction}, except that the commander, forgetfully or otherwise, excluded the conditional phrase, ``which has not been helped yet", about other inoperable robots to attempt rescues. In this case, if a rescue attempt fails (disabled robot remains disabled after attempted to be rescued), $R_i$ will go back to previous task {explore} ($T_1$), and immediately come back to the same \textit{rescue} task ($T_2$) again. Repeated failed rescue attempts will keep this cycle going on. Effectively, $R_i$ will keep attempting consecutive rescue attempts until $R_j$ is revived, and waste a lot of time. A more serious problem may occur if it gets disabled after making multiple failed rescue attempts, in case of a high-risk rescue. So, it might be worthwhile to detect any possibility of such situation so that the instruction can be corrected before robot $R_i$'s deployment.
 
 As we described, other robots getting disabled nearby as well as failed rescue attempts are the critical events that can cause this oscillation. These critical conditions are identified from the transition conditions on task transition model of robot $R_i$, between the two corresponding tasks. We perform detection of this situation with Monte-Carlo (MC) simulations and smart simulations, and compare the performances. The probability of critical events are estimated from the mission model. We use this example of low-probability situation to illustrate the value of smart simulations.

$20\text{,}000$ MC simulations give probability of oscillation to be $0.00015$. Therefore, it requires at least $14\text{,}000$ simulations to detect this situation confidently. However, with only $500$ simulations using our proposed framework -- a reduction in the \textit{required} number of simulations by a factor of at least $100$ -- we compute the probability of oscillation to be $0.00013$. 
Thus, smart simulation can provide an efficient way to approximate the estimated probability, and issue alerts.
 

\subsection {Using inferences to issue alerts}
\label {sec:inference}

We use arithmetic operations and graph search algorithms on state transition models to make quick inferences, and issue an alert instantly, when applicable.


\subsubsection{Is there a non-zero probability of Robot $R_i$ rescuing Robot $R_j$? } Let robot $R_i$ be in the field with after given the instruction ``Explore AoI-$1$ and AoI-$2$ sequentially, and rescue any disabled robot within the exploring AoIs. If event-$A$ happens, move to AoI-$3$, and explore AoI-$3$ and AoI-$4$ sequentially.'' Humans receive information on robot $R_j$ being inoperable in AoI-$2$, and $R_i$ is last known to be exploring AoI-$3$. We need to analyze whether $R_j$ will be attempted to be rescued by $R_i$. From a reachability test, the system may conclude in a moment that rescuing $R_j$ in AoI-$2$ is not a reachable state for $R_i$, given its latest known state. Any other robots, given their latest known states, do not have any \textit{relay} task among their reachable states, therefore no on-field robot's instruction-set will change. Therefore, it is guaranteed that $R_i$ will \textit{not} rescue $R_j$.


\section{Conclusions and Future Work}
\label{sec:conclusions}

We presented an alert generation framework for human-supervised multi-robot teams deployed in challenging applications. We identified several useful warning situations, and have expressed their conditions using Metric Temporal Logic (MTL) specification. Our state machine simulation with an abstract model of the mission is computationally-efficient, which is essential in HA/DR applications for rapid alert generation. More specifically, we probabilistically assessed complex conditions with forward simulations of the system and leveraged smart simulations to ensure a computationally-efficient way of detecting situations that have low probability. We also proposed an inference-based approach to detect some alert conditions with 
absolute certainty.

In the future, we would like to study the facets on the user end, like natural language processing for extracting the commands to a mathematical framework. We would also like to perform a human study with a well-designed interface, and work on identifying reasonable probability threshold values for the alert conditions by modelling a human user's preference from efficient choice experiments\cite{louviere2008modeling} on mission outcomes.



\textbf{Acknowledgment:}
\addcontentsline{toc}{section}{Acknowledgment}
This work was supported by the U.S. Army Research Laboratory. Any opinions or conclusions expressed in this paper are those of the authors and do not necessarily reflect the views of the sponsor.


\bibliographystyle{aaai}
\bibliography{main}

\begin{thebibliography}{}

\bibitem[\protect\citeauthoryear{Al-Hussaini, Gregory, and
  Gupta}{2018a}]{Al-Hussaini_2018_IROS}
Al-Hussaini, S.; Gregory, J.~M.; and Gupta, S.~K.
\newblock 2018a.
\newblock Generation of context-dependent policies for robot rescue
  decision-making in multi-robot teams.
\newblock In {\em IEEE/RSJ International Conference on Intelligent Robots and
  Systems (IROS)}.

\bibitem[\protect\citeauthoryear{Al-Hussaini, Gregory, and
  Gupta}{2018b}]{Al-Hussaini_2018_SSRR}
Al-Hussaini, S.; Gregory, J.~M.; and Gupta, S.~K.
\newblock 2018b.
\newblock A policy synthesis-based framework for robot rescue decision-making
  in multi-robot exploration of disaster sites.
\newblock In {\em IEEE International Symposium on Safety, Security and Rescue
  Robotics (SSRR)}.

\bibitem[\protect\citeauthoryear{Barnes \bgroup et al\mbox.\egroup
  }{2014}]{barnes2014designing}
Barnes, M.~J.; Chen, J.~Y.; Jentsch, F.; Oron-Gilad, T.; Redden, E.; Elliott,
  L.; and Evans~III, A.~W.
\newblock 2014.
\newblock Designing for humans in autonomous systems: Military applications.
\newblock Technical report, ARMY RESEARCH LAB ABERDEEN.

\bibitem[\protect\citeauthoryear{Chai \bgroup et al\mbox.\egroup
  }{2016}]{chai2016collaborative}
Chai, J.~Y.; Fang, R.; Liu, C.; and She, L.
\newblock 2016.
\newblock Collaborative language grounding toward situated human-robot
  dialogue.
\newblock {\em AI Magazine} 37(4):32--45.

\bibitem[\protect\citeauthoryear{Chien \bgroup et al\mbox.\egroup
  }{2018}]{chien2018attention}
Chien, S.-Y.; Lin, Y.-L.; Lee, P.-J.; Han, S.; Lewis, M.; and Sycara, K.
\newblock 2018.
\newblock Attention allocation for human multi-robot control: Cognitive
  analysis based on behavior data and hidden states.
\newblock {\em International Journal of Human-Computer Studies} 117:30--44.

\bibitem[\protect\citeauthoryear{DeCostanza \bgroup et al\mbox.\egroup
  }{2018}]{Decostanza2018}
DeCostanza, A.~H.; Marathe, A.~R.; Bohannon, A.; Evans, A.~W.; Palazzolo,
  E.~T.; Metcalfe, J.~S.; and McDowell, K.
\newblock 2018.
\newblock Enhancing human-agent teaming with individualized, adaptive
  technologies: A discussion of critical scientific questions.
\newblock Technical report, US Army Research Laboratory Aberdeen Proving Ground
  United States.

\bibitem[\protect\citeauthoryear{Diller \bgroup et al\mbox.\egroup
  }{2014}]{HFACS-medical-2014}
Diller, T.; Helmrich, G.; Dunning, S.; Cox, S.; Buchanan, A.; and Shappell, S.
\newblock 2014.
\newblock The human factors analysis classification system (hfacs) applied to
  health care.
\newblock {\em American Journal of Medical Quality} 29(3):181--190.
\newblock PMID: 23814026.

\bibitem[\protect\citeauthoryear{Gregory \bgroup et al\mbox.\egroup
  }{2016}]{Gregory2016c}
Gregory, J.; Fink, J.; Stump, E.; Twigg, J.; Rogers, J.; Baran, D.; Fung, N.;
  and Young, S.
\newblock 2016.
\newblock {Application of Multi-Robot Systems to Disaster-Relief Scenarios with
  Limited Communication}.
\newblock {\em Field and Service Robotics: Results of the 10th International
  Conference}  639--653.

\bibitem[\protect\citeauthoryear{Jentsch}{2016}]{jentsch2016human}
Jentsch, F.
\newblock 2016.
\newblock {\em Human-robot interactions in future military operations}.
\newblock CRC Press.

\bibitem[\protect\citeauthoryear{Konur}{2013}]{probabilistic_temporal_13}
Konur, S.
\newblock 2013.
\newblock A survey on temporal logics for specifying and verifying real-time
  systems.
\newblock {\em Frontiers of Computer Science} 7(3):370--403.

\bibitem[\protect\citeauthoryear{Kruijff \bgroup et al\mbox.\egroup
  }{2014}]{kruijff2014experience}
Kruijff, G.-J.~M.; Jan{\'\i}{\v{c}}ek, M.; Keshavdas, S.; Larochelle, B.;
  Zender, H.; Smets, N.~J.; Mioch, T.; Neerincx, M.~A.; Diggelen, J.; Colas,
  F.; et~al.
\newblock 2014.
\newblock Experience in system design for human-robot teaming in urban search
  and rescue.
\newblock In {\em Field and Service Robotics},  111--125.
\newblock Springer.

\bibitem[\protect\citeauthoryear{Levine and
  Williams}{2014}]{levine2014concurrent}
Levine, S.~J., and Williams, B.~C.
\newblock 2014.
\newblock Concurrent plan recognition and execution for human-robot teams.
\newblock In {\em Twenty-Fourth International Conference on Automated Planning
  and Scheduling}.

\bibitem[\protect\citeauthoryear{Louviere \bgroup et al\mbox.\egroup
  }{2008}]{louviere2008modeling}
Louviere, J.~J.; Street, D.; Burgess, L.; Wasi, N.; Islam, T.; and Marley,
  A.~A.
\newblock 2008.
\newblock Modeling the choices of individual decision-makers by combining
  efficient choice experiment designs with extra preference information.
\newblock {\em Journal of choice modelling} 1(1):128--164.

\bibitem[\protect\citeauthoryear{Makris \bgroup et al\mbox.\egroup
  }{2016}]{makris2016augmented}
Makris, S.; Karagiannis, P.; Koukas, S.; and Matthaiakis, A.-S.
\newblock 2016.
\newblock Augmented reality system for operator support in human--robot
  collaborative assembly.
\newblock {\em CIRP Annals} 65(1):61--64.

\bibitem[\protect\citeauthoryear{Mosier \bgroup et al\mbox.\egroup
  }{2017}]{mosier2017autonomous}
Mosier, K.~L.; Fischer, U.; Burian, B.~K.; and Kochan, J.~A.
\newblock 2017.
\newblock Autonomous, context-sensitive, task management systems and decision
  support tools i: Human-autonomy teaming fundamentals and state of the art.

\bibitem[\protect\citeauthoryear{Murphy}{2004}]{murphy2004human}
Murphy, R.~R.
\newblock 2004.
\newblock Human-robot interaction in rescue robotics.
\newblock {\em IEEE Transactions on Systems, Man, and Cybernetics, Part C
  (Applications and Reviews)} 34(2):138--153.

\bibitem[\protect\citeauthoryear{Murphy}{2014}]{murphy2014disaster}
Murphy, R.~R.
\newblock 2014.
\newblock {\em Disaster robotics}.
\newblock MIT press.

\bibitem[\protect\citeauthoryear{Ouaknine and Worrell}{2008}]{MTL_08}
Ouaknine, J., and Worrell, J.
\newblock 2008.
\newblock Some recent results in metric temporal logic.
\newblock In {\em International Conference on Formal Modeling and Analysis of
  Timed Systems},  1--13.
\newblock Springer.

\bibitem[\protect\citeauthoryear{Ramakrishnan, Zhang, and
  Shah}{2017}]{ramakrishnan2017perturbation}
Ramakrishnan, R.; Zhang, C.; and Shah, J.
\newblock 2017.
\newblock Perturbation training for human-robot teams.
\newblock {\em Journal of Artificial Intelligence Research} 59:495--541.

\bibitem[\protect\citeauthoryear{Ramchurn \bgroup et al\mbox.\egroup
  }{2016}]{ramchurn2016disaster}
Ramchurn, S.~D.; Huynh, T.~D.; Wu, F.; Ikuno, Y.; Flann, J.; Moreau, L.;
  Fischer, J.~E.; Jiang, W.; Rodden, T.; Simpson, E.; et~al.
\newblock 2016.
\newblock A disaster response system based on human-agent collectives.
\newblock {\em Journal of Artificial Intelligence Research} 57:661--708.

\bibitem[\protect\citeauthoryear{Rold{\'a}n \bgroup et al\mbox.\egroup
  }{2017}]{roldan2017multi}
Rold{\'a}n, J.; Pe{\~n}a-Tapia, E.; Mart{\'\i}n-Barrio, A.;
  Olivares-M{\'e}ndez, M.; Del~Cerro, J.; and Barrientos, A.
\newblock 2017.
\newblock Multi-robot interfaces and operator situational awareness: Study of
  the impact of immersion and prediction.
\newblock {\em Sensors} 17(8):1720.

\bibitem[\protect\citeauthoryear{Shappell \bgroup et al\mbox.\egroup
  }{2007}]{HFACS-aviation-2007}
Shappell, S.; Detwiler, C.; Holcomb, K.; Hackworth, C.; Boquet, A.; and
  Wiegmann, D.~A.
\newblock 2007.
\newblock Human error and commercial aviation accidents: An analysis using the
  human factors analysis and classification system.
\newblock {\em Human Factors} 49(2).

\bibitem[\protect\citeauthoryear{Sherwood}{2018}]{sherwood2018effect}
Sherwood, S.~M.
\newblock 2018.
\newblock The effect of task load, automation reliability, and environment
  complexity on uav supervisory control performance.

\bibitem[\protect\citeauthoryear{Shriyam and
  Gupta}{2018}]{shriyam2018incorporating}
Shriyam, S., and Gupta, S.~K.
\newblock 2018.
\newblock Incorporating potential contingency tasks in multi-robot mission
  planning.
\newblock In {\em IEEE International Conference on Robotics and Automation
  (ICRA)}.

\bibitem[\protect\citeauthoryear{Wong and Seet}{2017}]{wong2017workload}
Wong, C.~Y., and Seet, G.
\newblock 2017.
\newblock Workload, awareness and automation in multiple-robot supervision.
\newblock {\em International Journal of Advanced Robotic Systems}
  14(3):1729881417710463.

\end{thebibliography}

\end{document}